\documentclass[conference]{IEEEtran}
\usepackage{times}

\usepackage[numbers]{natbib}
\usepackage{multicol}
\usepackage[bookmarks=true]{hyperref}

\usepackage[pdftex]{graphicx}
\usepackage{color}
\usepackage{amssymb}
\usepackage{latexsym}
\usepackage{amsmath}
\usepackage{subfigure}

\begin{document}

\title{An Information Theoretic Measure for \\ Robot Expressivity}

\author{A. LaViers}



%

\maketitle

\begin{abstract}
This paper presents a principled way to think about articulated {movement} for artificial agents and a measurement of platforms that produce such movement.  In particular, in human-facing scenarios, the shape evolution of robotic platforms will become essential in creating systems that integrate -- and communicate -- with human counterparts.   This paper provides a tool to measure the \textit{expressive capacity} or \textit{expressivity} of articulated platforms.  To do this, it points to the synergistic relationship between computation and mechanization.  Importantly, this way of thinking gives an information theoretic basis for measuring and comparing robots of increasing complexity and capability.  The paper will  provide concrete examples of this measure in application to current robotic platforms.  It will also provide a comparison between the computational and mechanical capabilities of robotic platforms and analyze order-of-magnitude trends over the last 15 years.  Implications for future work made by the paper are to provide a method by which to quantify movement imitation, outline a way of thinking about designing expressive robotic systems, and contextualize the capabilities of current robotic systems. 
\end{abstract}
\IEEEpeerreviewmaketitle
\section{Introduction}
As robotic systems move outside of a factory and into human work places, public spaces, homes, and even bodies, the pattern of movement which each platform produces will become essential to understand and to design intentionally.  Further, as we aim to build and integrate multi-purpose robots that can adapt to many tasks and many scenarios, an understanding of how much a single platform can do -- relative to another -- will become important.  Further, in both of these spaces, a notion of human movement imitation is enticing, and thus, how success of such imitation is measured is important.  

This paper offers an approach to thinking about robotic platforms that can aid in each of these domains.  In particular, the idea of \textit{expressive movement} may guide how we measure the efficacy of robots in human-facing scenarios.  This is often described as the different manners a robot should move if it is in a care-giving setting versus an authoritarian one like a modulation of a superficial or decorative style that is independent of practical task. 

Robots that enter private spaces such as hospital rooms and homes should modulate their behavior in order to bring comfort and respect to human counterparts in such environments (just as humans modulate their own movement).  On the other hand, a robot guiding humans out of a burning building or directing traffic should use clear, aggressive movements to impart the gravity of an emergency situation and ensure that each motion conveys a clear command.  Thus, the ability to move \textit{expressively} increases the function of such platforms.  In these cases, we imagine that \textit{information} passes to the human counterpart based on the motions of the robotic platform.  

However, even in the factory setting, modulations can improve performance.  Take for example a robot that may need to modulate the motions it uses to apply paint to a surface in order to compensate for the thickness of the paint on a given day: sometimes a `flicking' quality will be more effective than a `dabbing' one.  Thus, to distinguish between functional and expressive movements is a bit of a matter of perspective. 

In either such cases (functional or expressive), we often attempt to \textit{imitate} the behavior of humans.  Indeed, it seems that in natural human settings the movement of humans encodes information.  To generalize the examples given here, this information may deal with environmental state (`is the building burning?'), task state (`how thick is the paint?'), or emotional state (`are you in a hurry?' and even `are you upset?').  Thus, we can think of an expressive robotic motion as an information {source} and the process of interpreting as a noisy channel to a receiver.  This paper will begin to formalize such a setup. 

Imitating the movement of biological organisms has been a topic in animation \cite{reynolds1999steering} and robotics \cite{breazeal2002robots,nakaoka2004leg,egerstedt2005ants,powell2012motion,ames2014human,knight2015layering}, which is often dependent on the parameterization of the creature's movement.   In \cite{yamane2004synthesizing,nakaoka2004leg,gillies2009learning,kulic2012incremental,laviers2014ICCPS,jiang2014analytic}  motion capture data is used to seed artificial representations. In several of these cases, continuous, trajectory-based measurements for success have been posed. The review in \cite{breazeal2002robots} discusses ``robots that imitate humans'' saying ``there are many ways in which a robot can be made to replicate the movement of a human'' and describing ``very high fidelity playback.''  

Instead, this paper takes the approach of measuring the complexity of such movement, via a discrete, information theoretic approach as \cite{restle1979coding,donald1995information} did for workspace sensing tasks.  In this case, we look to the dimensionality (rather than the trajectory recreation of an individual degree of freedom) of the data used to represent motion.    A review of datasets of human motion indicates this number is on the order of tens or hundreds of parameters \cite{sigal2010humaneva}.  Even in a new sensor-rich environment  \cite{joo2015panoptic}, models of a similar dimension are extracted.  In \cite{safonova2004synthesizing,chai2005performance} the notion of ``low-dimensional'' signals are introduced by using sparse (or, even sparser) marker sets; these representations are termed ``physically meaningful.''

When successful, a proper parameterization of movement reveals much about the biology of the animal as in  \cite{stephens07, stephens08a, stephens08b}.   In this work, the movement of a \textit{C. Elegan} was analyzed using a curve parameterized by 100 angles.  Then, after capturing many hours of behavior, a principle component analysis was performed, revealing that the structure of the behavior could be explained as a superposition of four primary poses \cite{stephens07}.  Observing the animal through this lens provided new insights into the behavior of this well-studied animal \cite{stephens08a,stephens08b}.  A \textit{C. Elegan} is a tiny worm with only 302 neurons governing its behavior.  The chosen 100 dimensional representation is then \textit{on the same order of magnitude as the number of the organism's neurons}.  
This parameterization is also on the same order of magnitude as the motion capture data of humans analyzed for robotics cited here.   However, given that there are thought to be 86 billion neurons in the human brain \cite{azevedo2009equal} (or certainly many more than in a \textit{C. Elegan}), it is likely that many, many more degrees of freedom are needed to represent -- or imitate -- human movement.  This motivates the need for another perspective with which to compare movement across platforms.

Thus, this paper presents an information theoretic measure of expressivity\footnote{Here, this term is used to describe capability of physical hardware, which is distinct from the use of the term in computer science to describe capability of a given coding language.} for robots in Section \ref{def}.  This measure is applied to several distinct robots in Section \ref{examples}.  Trends in the capacity for robots to exhibit expressive motion are analyzed using the measure in Section \ref{trends}; the measure reveals a possible gap between computation and mechanization in modern machines.  Discussion of a dynamic extension of the measure is provided in Section \ref{dynamic}.  The paper will offer concluding remarks in Section \ref{conclusion}.

\section{A Measure for Expressivity}
\label{def}
A formalism for the concept of \textit{expressivity} is provided here.  This definition is inspired from comparison of robots to Turing's notion of computation.  As has been motivated in the prior background section, the feature that very much differentiates verbose movement ``vocabularies'' in moving platforms (machines and animals) is the \textit{number of degrees of freedom needed to represent the movement and, thus, available to create complex movement}.  

Thus, we propose to track all possible configurations of the platform's degrees of freedom.  That is, we'll measure the size of the number of shapes that can be achieved kinematically, which is equivalent to the precision with which a number could be recorded on the platform (as transistors are used in computers).  We'll call this the  {\textit{kinematic mechanization capacity}}.  We contend that this can be seen as a \textit{{fundamental limit}} on the expressivity available to a robot.

To model this is simply a matter of capturing the representation with a number.  In analogy, computers use the unit of \textit{bits} to do this; more complex computational calculations require processors that can hold more bits, and thus use more transistors to do so.  Or, an 8-bit display is more \textit{expressive} than a 1-bit display.  Similarly, robots can be viewed as needing more mechanical configurations in order to complete more complex mechanizations.  A new mechanical configuration is created by a degree of freedom with more range of motion, more precision in its motion, or a new degree of freedom; any of these {\textit{increases the expressivity}} of a platform.

To formalize this idea, let $N$ be the number of actuator \textit{types} on a machine.  On computers with solid state hard drives, or other homogeneous machines, this number is $1$ since these machines are made up of many, many transistors.  For robots, we may often have most of the machine comprised of servo motors; however we may also have heterogeneity.  On a robot with a simple gripper (which is either open or closed) and two identical servos, $N=2$.  

Let $M_{i}$ be the number of degrees of freedom with a particular number of available configurations $R_i$, which is computed via counting from an actuators minimum to maximum range via its resolution where $i=1,...,N$.  For a robot comprised of two servos, say with $360^o$ range and $0.1^o$ resolution with a gripper that may be `open' or `closed', $R_1=3600$ with $M_1=2$ and $R_2=2$ with $M_2=1$.

From this description of a machine's construction, let
\begin{equation}
\mathcal{C}=\prod_{i=1}^N R_{i}^{M_i}
\end{equation}
be the number of geometric or kinematic configurations (shapes) available to a platform.  
For the simple robot with an open-close gripper and two servos this becomes $$2^1\times3600^2=2.6\times10^7 \mbox{ configurations.}$$  Another computation can compare this raw number of configurations to binary displays and computer architecture. 
From there the \textit{kinematic mechanization capacity} of that platform is
\vspace{-.1in}
\begin{equation}
\mathcal{K}=\log_2(\mathcal{C}).
\end{equation}
Thus, the robot in the previous example can express a shape containing $\approx 25\mbox{ bits}$ of information in its environment.  Then, just as a 25-bit display can represent $\approx 2.6 \times 10^7$ numbers, this simplistic robot can be compared to a 25-bit display.

In this section we present some toy examples:
\begin{itemize}
\item For a processor with 200 transistors: $N=1$, $M_1=200$, and $R_1=2$.  The quantity $\mathcal{K}$ is given by:
\begin{equation}
\log_2(2^{200}) = \log_2(1.6\times10^{60}\mbox{ configurations})=200\mbox{ bits}.\nonumber
\end{equation}
\item {For a manipulator with ten servos with a $0.1^o$ resolution and $360^o$ range of motion: $N=1$, $M_1=2$, and $R_1=3600$. The quantity $\mathcal{K}$ is given by:}
\begin{equation}
\log_2(3600^{10}) = \log_2(3.7\times10^{35}\mbox{ configurations})\approx118\mbox{ bits}.\nonumber
\end{equation}
\item {For the same manipulator with a gripper which can be open or closed: $N=2$, $M_1=2$, $R_1=3600$, $M_2=1$, $R_1=2$.   The quantity $\mathcal{K}$ is the same ($\approx118\mbox{ bits}$):}
\begin{eqnarray}
\log_2(3600^{10}\times2^1) = \log_2(3.7\times10^{35}\mbox{ configurations})\nonumber
\end{eqnarray}
\end{itemize}

\section{Applied Examples}
\label{examples}
In this section, the measure introduced in the previous section will be applied to some instructive examples.  In particular, we will compare a common humanoid robot and a machine which attracts many interested human viewers, Vegas's Bellagio fountains.

\subsection{Aldebaran NAO Humanoid Robot} \label{NAO_ex}
The Aldebaran NAO robot is a captivating machine.  Indeed, advances in feedback control and robotics were needed to build it.  Appropriately, there is something impressive about seeing it move.  The computation provided in this section, however, may call into question how useful it can be for replicating human behavior in any context.  

\begin{figure}[h!]
\centering
\includegraphics[width=\columnwidth]{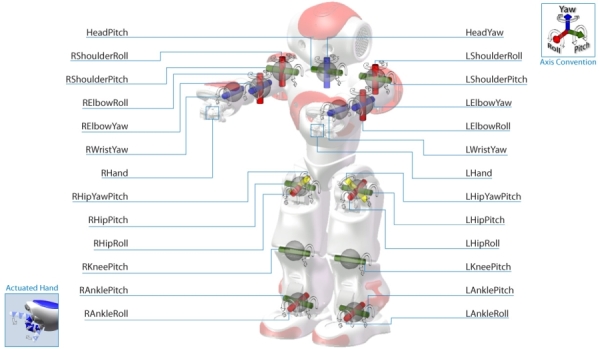}
\caption{A diagram which lists the degrees of freedom on an Aldebaran NAO robot.  In addition to these mechanical degrees of freedom the platform contains an ATOM Z530 onboard computer processor, which has 47 million transistors on board. \cite{G2009,aldebaran2009robotics}} 
\label{NAO}
\vspace{-.2in}
\end{figure}

Figure \ref{NAO} and Table \ref{nao_dof} outline the basic capabilities of the platform where the sensor resolution (an encoder with $0.1^o$ precision) has been used to determine $R_i$.

\begin{table}[h!] 
\begin{center}
\begin{tabular}{|p{.3\columnwidth}||p{.3\columnwidth}||p{.1\columnwidth}|}
\hline
{DOF} & {Range / Resolution} & \textbf{$R_i$}\\
  \hline \hline
    l/r hand & open/close & 2 \\
  \hline
    head yaw & -119.5 to 119.5 / .1 & 2390 \\
  \hline
    head pitch (at 0 yaw) & -38.5 to 29.5 / .1 & 680 \\
  \hline
    l/r shoulder pitch & -119.5 to 119.5 / .1 & 2390 \\
  \hline
      l/r shoulder yaw & -119.5 to 119.5 / .1 & 2390 \\
  \hline
      l/r shoulder roll & -88.5 to -2 / .1 & 865 \\
  \hline
      l/r wrist yaw & -104.5 to 104.5 / .1  & 2090 \\
  \hline
      pelvis & -65.6 to 42 / .1 & 1076 \\
  \hline
      l/r hip roll & -21.7 to 45.2 / .1 & 669 \\
  \hline
      l/r hip pitch & -88 to 27.7 / .1 & 1157 \\
  \hline
      l/r knee pitch & -5.3 to 121.0 / .1 & 1263 \\
  \hline
      l/r ankle pitch & -68.2 to 52.9 / .1 & 1211 \\
  \hline
      l/r ankle roll & -22.8 to 44.1 / .1 & 669 \\
  \hline
\end{tabular}
\end{center}
\caption{NAO Aldebaran robot, mechanical degrees of freedom.} \label{nao_dof}
\end{table}

Thus, the kinematic mechanization capacity is calculated as
\begin{eqnarray}
\mathcal{K}=\log_2(2^2\times2390^5\times680^1\times940^2\times865^2...\\\nonumber\times2090^2
\times1076^1\times669^4\times1157^2\times1263^2\times1211^2)\\\nonumber
=\log_2(4.1\times10^{71}\mbox{ configurations})\approx238 \mbox{ bits}
\end{eqnarray}

This calculation includes physical combinations which are kinematically or dynamically infeasible.   However, changes in motor speed between configurations could also increase the complexity perceived (see Section \ref{dynamic}), pointing out missed states. Thus, the number may be seen as an approximation.

\subsection{Bellagio Water Fountains} 
\label{bellagio_ex}
Consider a tourist attraction, like the Bellagio water fountains in Las Vegas, NV.  Tourists line up every hour to watch this famous display, routinely included in lists of popular Vegas attractions.  This is to say that the fountain display is visually very interesting, or expressive, for human watchers.  Let's compare how much more complex it is than typical robots and consider how much less complex it is than most computers via the proposed measure.

The fountain has about 1,200 water cannons with 5,000 lights as part of its display.  It also has the ability to create fog and features popular or famous music during the shows.  For this analysis \cite{vegas1,vegas2}, we'll consider only the water cannons and lights. The cannons come in four types: robotic Oarsman and three sizes of Shooters.  The 208 Oarsman are articulated cannons with active control; the Shooters simply blast water at three predetermined pressure settings, each having a single pressure setting according to their size.  The lights can be a range of colors.

\begin{table}[h!] 
\begin{center}
\begin{tabular}{|p{.35\columnwidth}||p{.3\columnwidth}||p{.1\columnwidth}|}
\hline
{DOF} & {Range / Resolution} & \textbf{$R_i$}\\
  \hline \hline
    Oarsmen RX (208) & $0^o$ to $160^o$ / by $1^o$ & 160 \\
  \hline
    Oarsmen RY (208) & $0^o$ to $160^o$ / by $1^o$ & 160 \\
  \hline
      Oarsmen water (208) & on/off & 2 \\
  \hline
Shooters (1,175) & on/off & 2\\
  \hline
lights (6,200) & off or one of 12 colors & 13 \\
  \hline
\end{tabular}
\end{center}
\caption{Estimated fountain system degrees of freedom.} \label{bellagio}
\vspace{-.2in}
\end{table}

Table \ref{bellagio} articulates a model for this system.  For the Oarsman, which rotate about two axes, we assume a range of motion of $160^o$ with a resolution of $1^o$ in each dimension.  We assume the water shooting out of the cannon to be on or off with a single pressure setting.  Likewise, the Shooters, are either on or off without articulation.  The lights can be `off' or one of twelve colors (as modeled by a moderate segmentation of the color wheel).  We ignore the music that plays alongside.


Thus, to compute the kinematic mechanization capacity, we find the following computation.
\vspace{-.05in}
\begin{eqnarray}
\mathcal{K}=\log_2(2^{1175+208}\times160^{208+208}\times13^{6200})\\\nonumber
=\log_2(4.9\times10^{8239}\mbox{ configurations})\\\nonumber
\approx27,372 \mbox{ bits}
\end{eqnarray}

We could argue over which is more interesting to watch: a NAO or the Bellagio fountains, but this metric provides a quantitative bound on \textit{how much more expressive} the fountains are.  In this case, about two orders of magnitude with respect to the amount of information they can encode.  This might strike roboticists as odd, but in terms of system expense and tourist attendance, the measure is consistent.

What if all the water cannons were the articulated, Oarsman variety?   In that case, the computation becomes:
\vspace{-.05in}
\begin{eqnarray}
\mathcal{K}=\log_2(2^{1383}\times160^{1383}\times13^{6200})\\\nonumber
=\log_2(1.2\times10^{10371}\mbox{ configurations})\\\nonumber
\approx34,452 \mbox{ bits}
\end{eqnarray}

Thus, we can see that by upgrading 1,175 cannons to the articulated variety, we don't gain much in expressivity.  If, in addition, we boost the resolution of each cannon of the original system to $0.1^o$, the following computation holds:

\begin{eqnarray}
\mathcal{K}=\log_2(2^{1383}\times1600^{1383}\times13^{6200})\\\nonumber
=\log_2(1.2\times10^{11754}\mbox{ configurations})\\\nonumber
\approx39,046\mbox{ bits}.
\end{eqnarray}

Thus here, we can see how adding water cannons and articulation resolution increases the expressivity of the platform, but we do not capture the additional expressivity that the dynamics of timing and water add to (and possibly take away from) the system.  For example, by moving with a certain timing, these fountains create different water displays, which add to the system's expressivity.  On the other hand, in the presence of water, moving with a particular force, not all points in the cannon's range might be physically feasible.  

\vspace{.1in}
\section{A Comparison Between Modern Computers and Modern Robots}
\label{trends}
This same measure has also been, previously, applied to computers.  A simple observation about the rate at which silicon chips were doubling their transistor count, dubbed Moore's Law, has been an important benchmark for computational capacity for the computer processor industry \cite{schaller1997moore}.  The premise of the importance of this observation is that more transistors offer more precision in number representation for a single computation.  Specifically, adding an additional transistor adds a \textit{bit} of capacity.  (Indeed, this is the origin of the unit used in the previous sections to measure robot expressive capacity.)

Robots typically have computers on board.  Increasing the computational power of such devices adds to the complexity of internal models for decision making.  
In this section, we'll compare robotic and computational hardware -- contained on the same platforms -- in order to point out order of magnitude trends between the computational and mechanical capacity of robots over the last decade and a half.

First, let us note that this distinction is a bit arbitrary.  Indeed, computer processors \textit{move}.  Due to the choice of transistors as effective computational elements, we don't see that movement expressed -- it's hidden within the electrons of tiny chips.  But, for many years computers were made out of mechanical elements.  We might, thus, see robots as a return to that original trend in computers.  In other words, robots are computers.  
Thus, in this section, we'll separate degrees of freedom dedicated to computation (transistors) from those dedicated to mechanization (motors), but future machines might meld the two, graying the distinction.

Finding details of many important robotic platforms over the last fifteen years has not been possible.  In order to provide a somewhat large sample size, the range of motion and precision of the actuators for most of the robots presented in this section have been estimated based on viewing motion of the machines and supplemented, where possible, with information from manufacturers and developers.  For most platforms, we assume $0.1^o$ actuator precision and estimate the range of motion from watching videos of the platforms' movement.  The computation presented in Section \ref{NAO_ex} is a representative example of those discussed here.  The full list of values used in this section is available in the Appendix.

\begin{figure}[h!]
\centering
\includegraphics[width=\columnwidth]{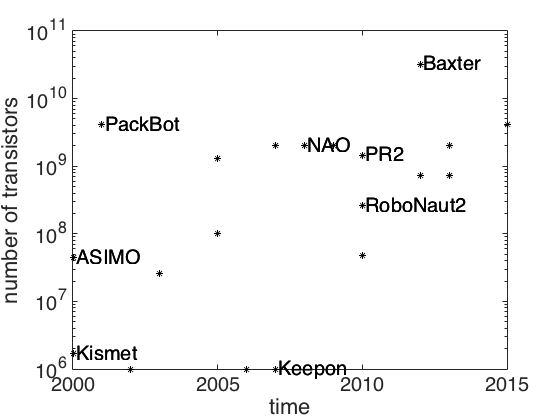}
\caption{The number of transistors used in computational processors of robots over the last fifteen years. Some platform names are omitted for clarity.}\label{fplot1}
\end{figure}

The plot in Figure \ref{fplot1} shows an analogous plot to those that revealed Moore's law, where the processors listed are housed in selected robotic platforms.  Plots like this have been used to track the progress of computational power over time, which has roughly doubled every year, even serving as a driving goal for the industry.  In Moore's plot, each additional component on an integrated circuit represents the ability to represent a larger -- or more precise -- number on a single chip and thus more precision with which to compute.  

Notably, every modern computer can perform many of the same operations, but this plot shows an increase in computational power, even onboard robotic platforms.  This representation occurs within the mechanism of transistors.  Each new transistor adds a new power of 2 in representation precision.  Note, that the \textit{the number of transistors} in modern, stand-alone processors is in the billions.  To convert that to \textit{{possible machine configurations}}, where actuators are transistors, the number 2 (which is the number of configurations for each actuator) has to be raised to that large number, resulting in a number of configurations that is on the order of $2^{10^{11}}$ or $10^{30102999566}$.

\begin{figure}[h!]
\centering
\includegraphics[width=\columnwidth]{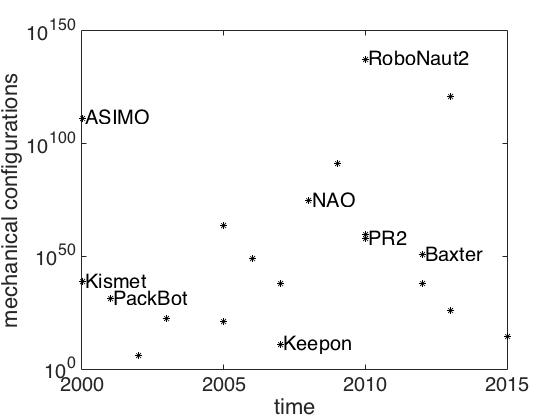}
\caption{The number of configurations available for mechanization in robot platforms over time.  Several platforms have been assumed to have a positioning resolution of $0.1^{o}$ and an estimated range of motion, and some names are omitted for clarity.  }
\label{fplot2}
\end{figure}

Figure \ref{fplot2} illuminates how the expressivity measure introduced in Section \ref{def} has evolved on robots over time.  Specifically, we plot number of kinematic configurations over time.  Like Moore's proxy of the number of transistors within a given CPU, this kinematic configuration space is not perfect -- dynamics will get in the way, as in a PC, if clock times aren't aligned and programming is inefficient, we can get better performance out of lower capacity machines -- but it gives a starting point for comparison.

\begin{figure}[h!]
\includegraphics[width=\columnwidth]{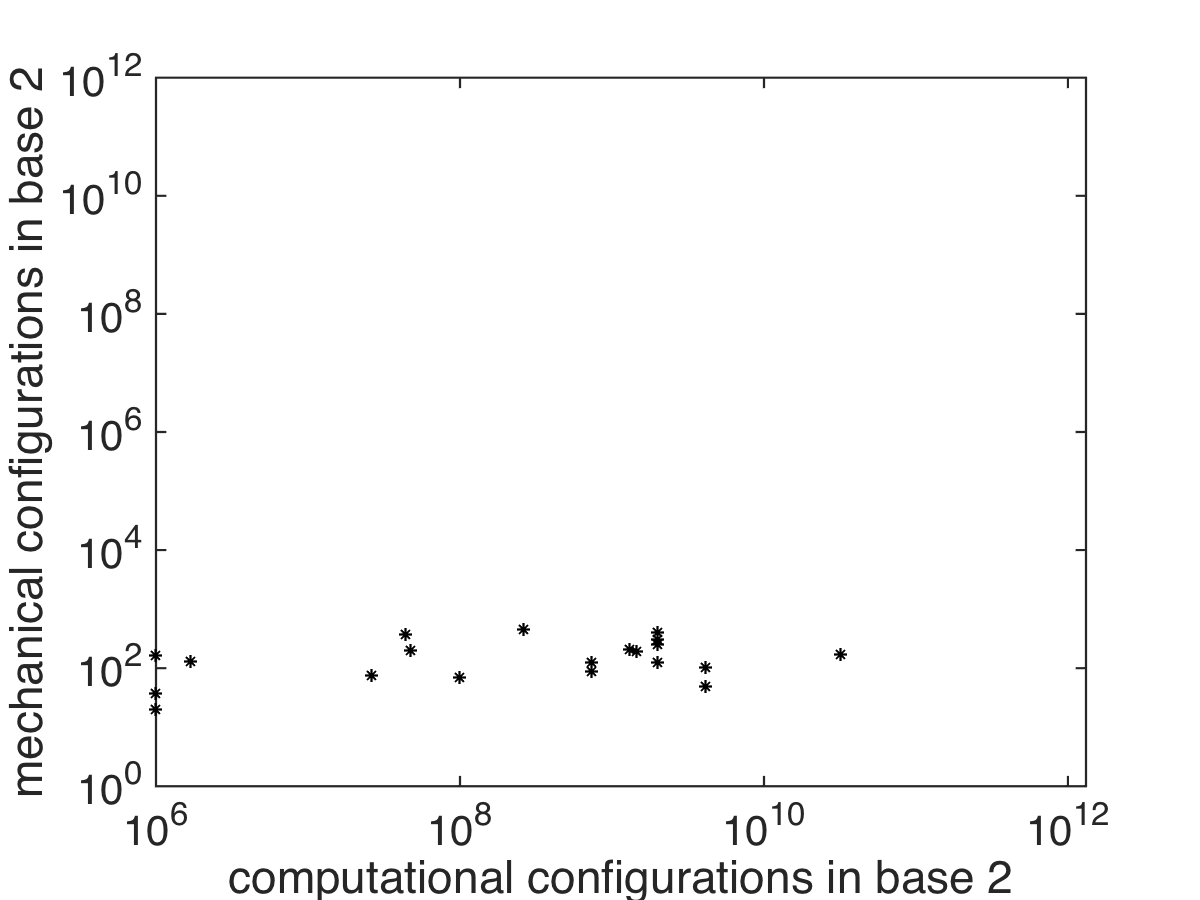}
\caption{A comparison of computational complexity relative to mechanical complexity on robotic platforms over time. Platform names are omitted for clarity.  The unit of measure on both axes is \textit{bits}.}
\label{fplot3}
\end{figure}

In parallel, by converting the number of configurations to a number represented in a base 2 number system, the rise in the computational capacities of these platforms can be compared to their mechanization capacities as in Figure \ref{fplot3}.  
This log-log plot provides a comparison in terms of the number of bits which it would take to describe the largest number that would fit in the onboard CPU versus the number of bits needed to represent each pose.  The plot shows a dramatic imbalance between computation and mechanization capacities.   

{{Consider, the NAO Aldebaran robot discussed in Section \ref{NAO_ex}.  It's kinematic configuration capacity is comparable to a 1960s computer chip with only 256 transistors}}.  The calculation in Section \ref{bellagio_ex} puts the famous fountain display on order of complexity of microprocessors made around 1980.  For example, the Intel 8086, which had 29,000 transistors.   

These initial plots are far from complete.  More analytical tools can supplement this analysis (see next section) as well as user studies for validation on how the biology of humans reacts to platforms.  However, the trends point to an interesting phenomenon, which has, anecdotally, surprised many roboticists.  Indeed, a prominent roboticist initially argued that an iPhone 6 has fewer available static configurations than the Baxter robot (something this analysis should put to rest).  Not only does the iPhone 6 have more configurations available, \textit{it has many, many orders of magnitude more}.  Further, this imbalance has remained flat over the past fifteen years.

This is an essential piece of information in the discussion \cite{stern2016raising,mcafee2016human,rankin2016basic}, which is by now mainstream \cite{ubi1,ubi2,ubi3}, around policy to support a rise in automation and the effects of job loss. The next section offers to develop a method to incorporate dynamic configuration strategies, which can both limit, in the case of physically infeasible poses, and expand, in the case of force sensing and force-controlled robots \cite{raibert1981hybrid,brogaardh2007present} the expressive capabilities of a particular platform.

\section{Toward Dynamic Expressive Capacity} \label{dynamic}
Firstly, note that we have over-approximated the shape space by ignoring dynamic effects and assuming that all shapes are dynamically feasible.  In this paper, every combination of actuator range of motion has been considered.  In real systems, however, not all of these shapes are kinematically or dynamically feasible.  For example, two robot arms may collide in certain cases of joint articulation, and this is a kinematically infeasible shape.  Additionally, some shapes may result in physical instabilities and cause the robot to fall over due to the effects of gravity; this is a dynamically infeasible shape.  For the order of magnitude analysis done here, it is unlikely that these infeasibiliities would greatly alter the results.

Further, two main sources of expressivity have been ignored here: additional configurations available to the source (to continue the analogy to communications) due to dynamic effects and inferences made over watching the source over time.  This section will spend some time discussing the former.  To the later, future work may investigate topics such as ``message size'' and ``information rate'' and ``channel capacity'' in the context of expressive robots based on the perspective presented here.  For example, it is likely that a good expressive robot will have some redundancy built into the system as humans are likely not perfect receivers (which may be modeled as a noisy channel).  

On the topic of additional configurations instigated by dynamics, we posit that the perceptual capabilities of humans need to be measured.  It's not clear how sensitive to velocity or force differentials humans are.  Indeed, in humans, dynamics are often coupled with explicit kinematic changes.  In particular, change in muscle tonus is a major source of interpretation of movement expression (and inference of inner state).  For example, a person sitting with a clenched jaw, visible through a bulge in their cheek, implies a different inner state than one without.  This expression is more viably modeled through the shape deformation of the person's cheek, rather than an explicit measurement of force within the mouth, which a human observer cannot directly intuit.  

Consider the example of the Bellagio fountains given in Section \ref{bellagio_ex} -- which clearly create different patterns in the water shot into the air based on the variable speed control used on the articulated cannons.  We know that for humans a timescale of about $100 ms$ is perceived to be instantaneous \cite{miller1968response}.  Thus, we can segment the states of the fountain machinery by $0.1 s$ to approximate the dynamic states.  Let's assume the Oarsman cannons have a maximum achievable angular velocity of $10^o/s$ or $100^o/100 ms$.

This assumption means that for each angle the articulated cannons can reach, there are $100$ additional states, corresponding to the different velocities at which the cannons can arrive to each position, which if in the `on' case where water is flowing, is visible to a human observer.  Thus, we have
\begin{eqnarray}
\log_2(2^{1383}\times(160\times100)^{416}\times13^{6200})\\\nonumber
=\log_2(1.2\times10^{13137}\mbox{ configurations})\\\nonumber
\approx43,640\mbox{ bits}.
\end{eqnarray}
This is of course does not account for the dynamics of the system in a formal way, but it further quantifies the concept of expressivity -- communicating information through movement -- of robotic systems.

\section{Conclusion} 
\label{conclusion}
This paper has presented a notion of expressive movement that shifts from a trajectory-based measure to a complexity-oriented one.  In particular, we draw a metaphor between machines for mechanization and computation.  In this way, the complexity of a mechanism on a machine describes its capability.  Here, the measure has analyzed order-of-magnitude capabilities of existing robotic platforms.

This measure firstly reveals a great dearth in motion imitation of biological systems.  Human motion in particular is a subject of robotics research and is often represented with 10s (order of magnitude) of degrees of freedom.  More sophisticated models in animation make it to 1,000s and 10,000s of parameters \cite{zordan2004breathe,sueda2008musculotendon}.  Yet, it took 100 joint angles to parameterize the motion of a simple \textit{C. Elegan}, which has only 302 neurons.  Thus, through the perspective presented here, the analysis of biological movement may offer an inroad to analyzing the complexity of animals, their brains, and the ability to create robotic systems that can mimic them.

Further, this measure has been used to show a stagnant trend in robotics: while computational resources have been heaped onto platforms, mechanical capacity has stayed on the same order-of-magnitude.  While computers have flourished due to the exponential growth of their capacity, robots have not focused on this area, favoring sophisticated control methods and computational decision models.  Indeed, factories with many robotic manipulators, may be viewed as systems which begin to show growth on the y-axis of the plot in Figure \ref{fplot3}.

Finally, the paper presents a formalization of the notion that the motion of robotic platforms can be used to \textit{communicate}.  Indeed many very low degree of freedom platforms have been used to do such a thing, with little regard to the fundamental capacity such platforms (with only translation and orientation at their disposal such as in the case of UAVs and UGVs) have for communication \cite{sharma2013communicating,knight2014expressive}.  
Note that these methods could also be used in multi-agent systems for robot-to-robot communication (although typically moving is more `expensive' than communicating for power-constrained systems).

Ongoing work is investigating extensions of this paper in a few distinct directions.  First, we are interest in how this can be used to formulate experiments in how humans perceive movement as in \cite{lee1976theory,restle1979coding,miller1968response} in the context of robotics.  Second, we are interested in how this method may inform architecture development, such as the use of motion primitives in teleoperation, where input versus output parameters of the system can be viewed as a compression channel (and, we hypothesize that good compression channels in this context will be better performing for users).  Finally, we are investigating how computation and mechanization may be specified for a machine in a synergistic manner to improve performance.

\section*{Acknowledgments}
This work was funded by DARPA award  \#D16AP00001.  Jialu Li and Varun Jain did essential work in helping to collect and estimate data on each platform reviewed.

\appendix
\section{Appendix} \label{appen}
Additional numbers used to calculate data points in Figures \ref{fplot1}-\ref{fplot3}.  For mechanical configurations, see below.  Many of these were determined through observation and are meant to indicate the visual expressiveness of the platforms.  Many rows represent multiple, homogeneous degrees of freedom (indicated through `l/r' for `left' and `right' and with numbers in parenthesis) for space.

\begin{table}[h!] 
\begin{center}
\begin{tabular}{|p{.32\columnwidth}||p{.3\columnwidth}||p{.08\columnwidth}|}
\hline
{DOF} & {Range / Resolution} & \textbf{$R_i$}\\
  \hline \hline
  Baxter &  &  \\
  \hline
  l/r S1 & open/close & 2 \\
  \hline
  l/r E1 & -2.864 to 150 / 0.1& 1530 \\
  \hline
    l/r W1 &-90 to 120 / 0.1& 2100 \\
  \hline
l/r S0 & -97.5 to 97.5 / 0.1 & 1950 \\
  \hline
l/r E0 &-175.0 to 175.0 / 0.1& 3500 \\
  \hline 
l/r W0 & -175.25 to 175.25 / 0.1 & 3505 \\
  \hline
l/r W2&-175.25 to 175.25 / 0.1 & 3505 \\
\hline
    Khepera IV & & \\
  \hline
l/r wheel & 360 / 0.1 / .1 & 3600 \\
  \hline
    Roomba & & \\
  \hline
l/r wheel & 360 / 0.1 / .1 & 3600 \\
  \hline
Kismet &  &  \\
  \hline
l/r ears pitch & -67.5 to 67.5 / 0.1  & 1350 \\
  \hline
l/r ears yaw & -22.5 to 22.5 / 0.1& 450 \\
  \hline
l/r eyelids &-1.5 to 1.5 / 0.1 & 30 \\
  \hline
l/r brows pitch & -10 to 10 / 0.1 & 200 \\
  \hline
l/r lips & -30 to 30 / 0.1 & 600 \\
  \hline
jaw & -22.5 to 22.5 / 0.1& 450 \\
  \hline
      \end{tabular}
\end{center}
\end{table}

\begin{table}[h!] 
\begin{center}
\begin{tabular}{|p{.32\columnwidth}||p{.3\columnwidth}||p{.08\columnwidth}|}
\hline
{DOF} & {Range / Resolution} & \textbf{$R_i$}\\
  \hline \hline
PR2 & & \\
  \hline
l/r shoulder pan& 170 / 0.1& 1700 \\
  \hline
l/r shoulder tilt&115 / 0.1 & 1150 \\
  \hline
l/r upper arm roll& 270 / 0.1 & 2700 \\
  \hline
l/r elbow flex &140 / 0.1 & 1400 \\
  \hline
l/r forearm roll & 360 / 0.1& 3600 \\
  \hline
l/r wrist pitch &130 / 0.1& 1300 \\
  \hline
l/r wrist roll&360 / 0.1  & 3600 \\
  \hline
head pan& 350 / 0.1& 3500 \\
  \hline
head tilt&115 / 0.1& 1150 \\
  \hline
Big Dog &  & \\
  \hline
each leg (5) (x4) & 150 / 0.08 & 1875 \\
  \hline
ASIMO & & \\
  \hline
head (3)& 150 / 0.08& 1875 \\
  \hline    
arms (14) & 150 / 0.08& 1875 \\
  \hline
hands (4)&150 / 0.08 & 1875 \\
  \hline
torso (1) & 150 / 0.08& 1875 \\
  \hline
legs (12) & 150 / 0.08& 1875 \\
  \hline
Little Dog & &\\
  \hline
l/r front knee RY&-177 to 57 / 0.1& 2340 \\
  \hline
l/r front hip RX& -34 to 34 / 0.1  & 680 \\
  \hline
l/r front hip RY& -200 to 137 / 0.1& 337 \\
  \hline
l/r back knee RY&-57 to177 / 0.1 & 2340 \\
  \hline
l/r back hip RX & -34 to 34 / 0.1 & 680 \\
  \hline
l/r back hip RY & -137 to 200 / 0.1 & 337 \\
  \hline
Robotnaut2 & & \\
  \hline
head yaw/pitch/roll  &150 / 0.08 & 1875 \\
  \hline
l/r hands (12) & 150 / 0.08 & 1875 \\
  \hline
l/r arms (7) & 150 / 0.08 & 1875 \\
  \hline
KeepOn &  &  \\
  \hline
tilt &-40 to 40 / 0.08& 1000 \\
  \hline
pan& -180 to 180 / 0.08 & 4500 \\
  \hline
pon & 0 to 100 / 0.08 & 1250 \\
  \hline
side & -25 to 25 / 0.08 & 625 \\
  \hline
RoboSapien&  &  \\
  \hline
l/r elbows & -90 to 90 / 0.1 & 1800 \\
  \hline
l/r shoulders & -30 to 150 / 0.1 & 1800 \\
  \hline
torso & -67.5 to 67.5 / 0.1 & 1350 \\
  \hline
l/r hips & -60 to 60 / 0.1 & 1200 \\
        \hline
Darwin &  &  \\
  \hline
neck pitch& -25 to 25 / 0.1 & 500 \\
  \hline    
neck roll& -90 to 90 / 0.1 & 1800 \\
  \hline
l/r elbow & 0 to 150 / 0.1 & 1500 \\
  \hline
l/r shoulder rotation & -100 to 100 / 0.1 & 2000 \\
  \hline
l/r shoulder compression & -15 to 15 / 0.1 & 300 \\
  \hline
l/r knee & 0 to 150 / 0.1 & 1500 \\
  \hline
l/r foot& 0 to 90 / 0.1 & 900 \\
  \hline
l/r waist rotation & -15 to 15 / 0.1  & 300 \\
  \hline
l/r knee/foot & -75 to 75 / 0.1 & 1500 \\
  \hline
l/r waist bend & 0 to 100 / 0.1 & 1000 \\
  \hline
\end{tabular}
\end{center}
\end{table}

\begin{table}[h!] 
\begin{center}
\begin{tabular}{|p{.32\columnwidth}||p{.3\columnwidth}||p{.08\columnwidth}|}
\hline
{DOF} & {Range / Resolution} & \textbf{$R_i$}\\
  \hline \hline
  Aibo &  &  \\
  \hline
head pan & -89 to 89 / 0.1 & 1780 \\
  \hline
head tilt& -62.5 to 62.5 / 0.1& 1250 \\
  \hline
head roll& -29 to 29 / 0.1 & 580 \\
        \hline
shoulders (4) & 0 to 100 / 0.1 & 1000 \\
  \hline
torso & -117 to 117 / 0.1& 2340 \\
  \hline    
knees (4)& 0 to 175 / 0.1 & 1750 \\
  \hline
l/r ears &0 to 20 / 0.1 & 200 \\
  \hline
tail (front to back) & -22.5 to 22.5 / 0.1 & 450 \\
  \hline
tail (left to right) & -12.5 to 12.5 / 0.1 & 250 \\
  \hline
Packbot &  &  \\
  \hline
shoulder rot. & 0 to 360 / 0.1& 3600 \\
  \hline
shoulder pivot&0 to 220 / 0.1& 2200 \\
  \hline
E1 pivot & 0 to 340 / 0.1& 3400 \\
  \hline
E2 pivot& 0 to 340 / 0.1& 3400 \\
  \hline
gripper rot.&0 to 360 / 0.1& 3600 \\
  \hline
gripper I/O & 180 / 0.1  & 1800 \\
  \hline
head rot. & 0 to 360 / 0.1& 3600 \\
  \hline
flipper& 0 to 360 / 0.1& 3600 \\
  \hline
Simon & &\\
  \hline
torso (2) &-75 to 75 / 0.1 & 1500 \\
  \hline
l/r arm (7) & 0 to 200 / 0.1 & 2000 \\
  \hline
face (5)&0 to 200 / 0.1 & 2000 \\
  \hline
Cheetah & &\\
  \hline
hip rot. (4)& 0 to 30 / 0.1 & 300 \\
  \hline
  hip (4)& 0 to 150 / 0.1& 1500 \\
  \hline
knee (4) & 0 to 200 / 0.1 & 2000 \\
  \hline
spine& -10 to 10 / 0.1 & 200 \\
  \hline
LBR iiwa & &\\
  \hline
axis 1& -170 to 170 / 0.1& 3400 \\
  \hline
axis 2& -120 to 120 / 0.1& 2400 \\
  \hline
axis 3& -170 to 170 / 0.1& 3400 \\
  \hline
axis 4& -120 to 120 / 0.1& 2400 \\
  \hline
axis 5& -170 to 170 / 0.1& 3400 \\
  \hline
axis 6& -120 to 120 / 0.1& 2400 \\
  \hline
axis 7& -175 to 175 / 0.1& 3500 \\
  \hline
KR60HA& & \\
  \hline    
axis 1& -185 to 185 / 0.1& 3700 \\
  \hline
axis 2&-135 to 35 / 0.1& 1700 \\
  \hline
axis 3& -120 to 158 / 0.1& 1780 \\
  \hline
axis 4& -350 to 350 / 0.1& 7000 \\
  \hline
axis 5& -119 to 119 / 0.1& 2380 \\
  \hline
axis 6& -350 to 350 / 0.1& 7000 \\
  \hline
\end{tabular}
\end{center}
\end{table}

For computational configurations, the following values were used. Here, $\mathcal{C}$ is $2^x$ where $x$ is the number of transistors.  Indeed, often, another, larger computer (or cluster of processors) is networked to these machines through wireless or wired connections.  But, it is instructive nonetheless to compare how much more sophisticated the computational power (even that which is on board) is relative to the mechanical power.

\begin{table}[h!] 
\begin{center}
\begin{tabular}{|p{.2\columnwidth}||p{.3\columnwidth}||p{.2\columnwidth}|}
\hline
Robot & Processor & {\# of transistors}\\
  \hline \hline
NAO &   & \\
  \hline
Baxter &  3rd Gen Intel Core i7-3770  & 1.40E+09\\
  \hline
      Khepera IV &  ARM Cortex-A8  & 2.00E+09\\
  \hline
    Roomba & & 1.00E+06\\
  \hline
Kismet & Motorola 68332 (4) &  1.68E+06\\
  \hline
PR2 & Two Quad-Core i7 Xeon  (8 cores) & 1.462E+09 \\
  \hline
Big Dog & Pentium CPU & 1.30E+09\\
  \hline
ASIMO & Pentium III-M 1.2 GHz & 4.40E+07\\
  \hline
Little Dog & Pentium CPU & 2.00E+09\\
  \hline
Robotnaut2 & & 2.622E+08 \\
  \hline
KeepOn & PS234 & 1.00E+06 \\
  \hline
RoboSapien& 200MHz ARM9 & 2.60E+07 \\
  \hline
Darwin & Intel Atom Z510 & 4.70E+07 \\
  \hline
Aibo & 64 bit RISC  & 1.00E+06 \\
  \hline
Packbot & Pentium 3 & 4.50E+07 \\
  \hline
Simon & & 2.00E+09\\
  \hline
Cheetah & & 7.31E+08\\
  \hline
LBR iiwa & & 7.31E+08\\
  \hline
KR60HA& & 1.00E+08\\
  \hline    
\end{tabular}
\end{center}
\end{table}

\newpage
\bibliographystyle{plainnat}
\bibliography{papers}

\begin{thebibliography}{41}
\providecommand{\natexlab}[1]{#1}
\providecommand{\url}[1]{\texttt{#1}}
\expandafter\ifx\csname urlstyle\endcsname\relax
  \providecommand{\doi}[1]{doi: #1}\else
  \providecommand{\doi}{doi: \begingroup \urlstyle{rm}\Url}\fi

\bibitem[Aldebaran()]{aldebaran2009robotics}
NAO Aldebaran.
\newblock {Robotics http://www. aldebaran-robotics. com/eng}.

\bibitem[Ames(2014)]{ames2014human}
Aaron~D Ames.
\newblock Human-inspired control of bipedal walking robots.
\newblock \emph{IEEE Transactions on Automatic Control}, 59\penalty0
  (5):\penalty0 1115--1130, 2014.

\bibitem[Azevedo et~al.(2009)Azevedo, Carvalho, Grinberg, Farfel, Ferretti,
  Leite, Lent, and Herculano-Houzel]{azevedo2009equal}
Frederico~AC Azevedo, Ludmila~RB Carvalho, Lea~T Grinberg, Jos{\'e}~Marcelo
  Farfel, Renata~EL Ferretti, Renata~EP Leite, Roberto Lent, and Suzana
  Herculano-Houzel.
\newblock Equal numbers of neuronal and nonneuronal cells make the human brain
  an isometrically scaled-up primate brain.
\newblock \emph{Journal of Comparative Neurology}, 513\penalty0 (5):\penalty0
  532--541, 2009.

\bibitem[Breazeal and Scassellati(2002)]{breazeal2002robots}
Cynthia Breazeal and Brian Scassellati.
\newblock Robots that imitate humans.
\newblock \emph{Trends in cognitive sciences}, 6\penalty0 (11):\penalty0
  481--487, 2002.

\bibitem[Brog{\aa}rdh(2007)]{brogaardh2007present}
Torgny Brog{\aa}rdh.
\newblock Present and future robot control development---an industrial
  perspective.
\newblock \emph{Annual Reviews in Control}, 31\penalty0 (1):\penalty0 69--79,
  2007.

\bibitem[Chai and Hodgins(2005)]{chai2005performance}
Jinxiang Chai and Jessica~K Hodgins.
\newblock Performance animation from low-dimensional control signals.
\newblock In \emph{ACM Transactions on Graphics (TOG)}, volume~24, pages
  686--696. ACM, 2005.

\bibitem[Donald(1995)]{donald1995information}
Bruce~Randall Donald.
\newblock On information invariants in robotics.
\newblock \emph{Artificial Intelligence}, 72\penalty0 (1-2):\penalty0 217--304,
  1995.

\bibitem[Echelon()]{vegas1}
Echelon.
\newblock Choreographed control.

\bibitem[Gillies(2009)]{gillies2009learning}
Marco Gillies.
\newblock Learning finite-state machine controllers from motion capture data.
\newblock \emph{Computational Intelligence and AI in Games, IEEE Transactions
  on}, 1\penalty0 (1):\penalty0 63--72, 2009.

\bibitem[Gouaillier et~al.(2009)Gouaillier, Hugel, Blazevic, Kilner, Monceaux,
  Lafourcade, Marnier, Serre, and Maisonnier]{G2009}
David Gouaillier, Vincent Hugel, Pierre Blazevic, Chris Kilner, Jerome
  Monceaux, Pascal Lafourcade, Brice Marnier, Julien Serre, and Bruno
  Maisonnier.
\newblock Mechatronic design of nao humanoid.
\newblock pages 769--774, National Institute of Aerospace, Jun 2009. Robotics
  and Automation, 2009. IEEE International Conference on.

\bibitem[Greg~Stephens and Bialek(2010)]{stephens08b}
Will~Ryu Greg~Stephens and William Bialek.
\newblock {The emergence of stereotyped behaviors in C. elegans}.
\newblock \emph{Bulletin of the American Physical Society}, 55, 2010.

\bibitem[Greg~Stephens and Ryu(2008)]{stephens07}
William~Bialek Greg~Stephens, Bethany Johnson-Kerner and Will Ryu.
\newblock {Dimensionality and dynamics in the behavior of C. elegans}.
\newblock \emph{PLoS Comput Biol}, 4\penalty0 (4):\penalty0 e1000028, 2008.

\bibitem[Greg~Stephens and Warrant(2010)]{stephens08a}
William Bialek Will~Ryu Greg~Stephens, Bethany Johnson-Kerner and E.~Warrant.
\newblock {From Modes to Movement in the Behavior of Caenorhabditis elegans}.
\newblock \emph{PloS one}, 5\penalty0 (11):\penalty0 462--465, 2010.
\newblock ISSN 1932-6203.

\bibitem[Jiang et~al.(2014)Jiang, Duerstock, and Wachs]{jiang2014analytic}
Hairong Jiang, Bradley~S Duerstock, and Juan~P Wachs.
\newblock An analytic approach to decipher usable gestures for quadriplegic
  users.
\newblock In \emph{Systems, Man and Cybernetics (SMC), 2014 IEEE International
  Conference on}, pages 3912--3917. IEEE, 2014.

\bibitem[Joo et~al.(2015)Joo, Liu, Tan, Gui, Nabbe, Matthews, Kanade, Nobuhara,
  and Sheikh]{joo2015panoptic}
Hanbyul Joo, Hao Liu, Lei Tan, Lin Gui, Bart Nabbe, Iain Matthews, Takeo
  Kanade, Shohei Nobuhara, and Yaser Sheikh.
\newblock Panoptic studio: A massively multiview system for social motion
  capture.
\newblock In \emph{Proceedings of the IEEE International Conference on Computer
  Vision}, pages 3334--3342, 2015.

\bibitem[Journal()]{vegas2}
Las Vegas~Review Journal.
\newblock Fountains of bellagio continue to amaze visitors.

\bibitem[Knight and Simmons(2014)]{knight2014expressive}
Heather Knight and Reid Simmons.
\newblock Expressive motion with x, y and theta: Laban effort features for
  mobile robots.
\newblock In \emph{The 23rd IEEE International Symposium on Robot and Human
  Interactive Communication}, pages 267--273. IEEE, 2014.

\bibitem[Knight and Simmons(2015)]{knight2015layering}
Heather Knight and Reid Simmons.
\newblock Layering laban effort features on robot task motions.
\newblock In \emph{Proceedings of the Tenth Annual ACM/IEEE International
  Conference on Human-Robot Interaction Extended Abstracts}, pages 135--136.
  ACM, 2015.

\bibitem[Kuli{\'c} et~al.(2012)Kuli{\'c}, Ott, Lee, Ishikawa, and
  Nakamura]{kulic2012incremental}
Dana Kuli{\'c}, Christian Ott, Dongheui Lee, Junichi Ishikawa, and Yoshihiko
  Nakamura.
\newblock Incremental learning of full body motion primitives and their
  sequencing through human motion observation.
\newblock \emph{The International Journal of Robotics Research}, 31\penalty0
  (3):\penalty0 330--345, 2012.

\bibitem[LaViers and Egerstedt(2014)]{laviers2014ICCPS}
Amy LaViers and Magnus Egerstedt.
\newblock Style-based abstractions for human motion classification.
\newblock \emph{ACM/IEEE 5th International Conference on Cyber-Physical
  Systems}, 2014.

\bibitem[Lee(1976)]{lee1976theory}
David~N Lee.
\newblock A theory of visual control of braking based on information about
  time-to-collision.
\newblock \emph{Perception}, 5\penalty0 (4):\penalty0 437--459, 1976.

\bibitem[Magnus~Egerstedt and Khan(2005)]{egerstedt2005ants}
Frank Dellaert Florent~Delmotte Magnus~Egerstedt, Tucker~Balch and Zia Khan.
\newblock What are the ants doing? vision-based tracking and reconstruction of
  control programs.
\newblock In \emph{Proceedings of the IEEE International Conference on Robotics
  and Automation (ICRA 2005)}, pages 18--22, 2005.

\bibitem[Manjoo(March 2, 2016)]{ubi1}
Farhad Manjoo.
\newblock A plan in case robots take the jobs: Give everyone a paycheck, March
  2, 2016.

\bibitem[McAfee and Brynjolfsson(2016)]{mcafee2016human}
Andrew McAfee and Erik Brynjolfsson.
\newblock Human work in the robotic future: Policy for the age of automation.
\newblock \emph{Foreign Affairs}, 95\penalty0 (4):\penalty0 139, 2016.

\bibitem[Miller(1968)]{miller1968response}
Robert~B Miller.
\newblock Response time in man-computer conversational transactions.
\newblock In \emph{Proceedings of the December 9-11, 1968, fall joint computer
  conference, part I}, pages 267--277. ACM, 1968.

\bibitem[Murray(June 3, 2016)]{ubi3}
Charles Murray.
\newblock A guaranteed income for every american, June 3, 2016.

\bibitem[Nakaoka et~al.(2004)Nakaoka, Nakazawa, Yokoi, and
  Ikeuchi]{nakaoka2004leg}
Shinichiro Nakaoka, Atsushi Nakazawa, Kazuhito Yokoi, and Katsushi Ikeuchi.
\newblock Leg motion primitives for a dancing humanoid robot.
\newblock In \emph{Robotics and Automation, 2004. Proceedings. ICRA'04. 2004
  IEEE International Conference on}, volume~1, pages 610--615. IEEE, 2004.

\bibitem[Porter and Manjoo(March 8, 2016)]{ubi2}
Eduardo Porter and Farhad Manjoo.
\newblock A future without jobs? two views of the changing work force, March 8,
  2016.

\bibitem[Powell et~al.(2012)Powell, Zhao, and Ames]{powell2012motion}
Matthew~J Powell, Huihua Zhao, and Aaron~D Ames.
\newblock Motion primitives for human-inspired bipedal robotic locomotion:
  walking and stair climbing.
\newblock In \emph{Robotics and Automation (ICRA), 2012 IEEE International
  Conference on}, pages 543--549. IEEE, 2012.

\bibitem[Raibert and Craig(1981)]{raibert1981hybrid}
Marc~H Raibert and John~J Craig.
\newblock Hybrid position/force control of manipulators.
\newblock \emph{Journal of Dynamic Systems, Measurement, and Control},
  103\penalty0 (2):\penalty0 126--133, 1981.

\bibitem[Rankin(2016)]{rankin2016basic}
Keith Rankin.
\newblock Basic income as public equity: The new zealand case.
\newblock In \emph{Basic Income in Australia and New Zealand}, pages 29--51.
  Springer, 2016.

\bibitem[Restle(1979)]{restle1979coding}
Frank Restle.
\newblock Coding theory of the perception of motion configurations.
\newblock \emph{Psychological Review}, 86\penalty0 (1):\penalty0 1, 1979.

\bibitem[Reynolds(1999)]{reynolds1999steering}
Craig~W Reynolds.
\newblock Steering behaviors for autonomous characters.
\newblock In \emph{Game developers conference}, volume 1999, pages 763--782,
  1999.

\bibitem[Safonova et~al.(2004)Safonova, Hodgins, and
  Pollard]{safonova2004synthesizing}
Alla Safonova, Jessica~K Hodgins, and Nancy~S Pollard.
\newblock Synthesizing physically realistic human motion in low-dimensional,
  behavior-specific spaces.
\newblock \emph{ACM Transactions on Graphics (TOG)}, 23\penalty0 (3):\penalty0
  514--521, 2004.

\bibitem[Schaller(1997)]{schaller1997moore}
Robert~R Schaller.
\newblock Moore's law: past, present and future.
\newblock \emph{IEEE spectrum}, 34\penalty0 (6):\penalty0 52--59, 1997.

\bibitem[Sharma et~al.(2013)Sharma, Hildebrandt, Newman, Young, and
  Eskicioglu]{sharma2013communicating}
Megha Sharma, Dale Hildebrandt, Gem Newman, James~E Young, and Rasit
  Eskicioglu.
\newblock Communicating affect via flight path exploring use of the laban
  effort system for designing affective locomotion paths.
\newblock In \emph{Human-Robot Interaction (HRI), 2013 8th ACM/IEEE
  International Conference on}, pages 293--300. IEEE, 2013.

\bibitem[Sigal et~al.(2010)Sigal, Balan, and Black]{sigal2010humaneva}
Leonid Sigal, Alexandru~O Balan, and Michael~J Black.
\newblock Humaneva: Synchronized video and motion capture dataset and baseline
  algorithm for evaluation of articulated human motion.
\newblock \emph{International journal of computer vision}, 87\penalty0
  (1-2):\penalty0 4, 2010.

\bibitem[Stern(2016)]{stern2016raising}
Andy Stern.
\newblock \emph{Raising the Floor: How a Universal Basic Income Can Renew Our
  Economy and Rebuild the American Dream}.
\newblock PublicAffairs, 2016.

\bibitem[Sueda et~al.(2008)Sueda, Kaufman, and Pai]{sueda2008musculotendon}
Shinjiro Sueda, Andrew Kaufman, and Dinesh~K Pai.
\newblock Musculotendon simulation for hand animation.
\newblock In \emph{ACM Transactions on Graphics (TOG)}, volume~27, page~83.
  ACM, 2008.

\bibitem[Yamane et~al.(2004)Yamane, Kuffner, and
  Hodgins]{yamane2004synthesizing}
Katsu Yamane, James~J Kuffner, and Jessica~K Hodgins.
\newblock Synthesizing animations of human manipulation tasks.
\newblock In \emph{ACM Transactions on Graphics (TOG)}, volume~23, pages
  532--539. ACM, 2004.

\bibitem[Zordan et~al.(2004)Zordan, Celly, Chiu, and
  DiLorenzo]{zordan2004breathe}
Victor~Brian Zordan, Bhrigu Celly, Bill Chiu, and Paul~C DiLorenzo.
\newblock Breathe easy: model and control of simulated respiration for
  animation.
\newblock In \emph{Proceedings of the 2004 ACM SIGGRAPH/Eurographics symposium
  on Computer animation}, pages 29--37. Eurographics Association, 2004.

\end{thebibliography}

\end{document}